# Towards Co-operative Congestion Mitigation

Aamir Hasan, Neeloy Chakraborty, Cathy Wu and Katherine Driggs-Campbell

*Abstract*— The effects of traffic congestion are widespread and are an impedance to everyday life. Piecewise constant driving policies have shown promise in helping mitigate traffic congestion in simulation environments. However, no works currently test these policies in situations involving real human users. Thus, we propose to evaluate these policies through the use of a shared control framework in a collaborative experiment with the human driver and the driving policy aiming to co-operatively mitigate congestion. We intend to use the CARLA simulator alongside the Flow framework to conduct user studies to evaluate the affect of piecewise constant driving policies. As such, we present our in-progress work in building our framework and discuss our proposed plan on evaluating this framework through a human-in-the-loop simulation user study.

## I. INTRODUCTION

Traffic congestion has an immense negative impact on society by affecting various facets such as urban mobility, climate change, and the economy [1]–[3]. Deploying a few autonomous vehicles on the road in idealized traffic settings has shown promise in eliminating congestion by improving the average speeds of vehicles [4]. However, the robust fully autonomous vehicles (AVs) required by such methods are unlikely to be available in the near future. Fortunately, shared control schemes through advanced driver assistive systems have shown that humans following instructions by smarter algorithms are viable stand-ins until robust autonomous vehicles are on the road [5].

Prior works have shown that a single Reinforcement Learning (RL) controlled autonomous agent can help stabilize traffic flow and stop the formation of traffic waves in the environment [4], [6]. Studies have also shown that simple speed management techniques can be used to improve both safety and emissions [2], [7]. Motivated by these factors and the human-compatible nature of piecewise constant policies, Sridhar and Wu proposed the use of 'Piecewise Constant Policies for Human-Compatible Congestion Mitigation' [8].

Our work is a direct extension to the framework proposed by Sridhar and Wu. In their paper, the authors describe policies that provide periodic "advice" to human drivers to modify their driving behaviour and mitigate congestion [8]. The policies are said to be piecewise constant as the action is expected to be held for ∆ timesteps in order to facilitate better adoption by human drivers. Though these policies were more robust and showed improvements on extension

A. Hasan, N. Chakraborty, and K. Driggs-Campbell are with the Department of Electrical and Computer Engineering at the University of Illinois at Urbana-Champaign. C. Wu is with the MIT Laboratory for Information & Decision Systems (LIDS), the Department of Civil and Environmental Engineering (CEE), and the Institute for Data, Systems, & Society (IDSS) at the Massachusetts Institute of Technology . Emails: {aamirh2, neeloyc2, krdc}@illinois.edu and cathywu@mit.edu

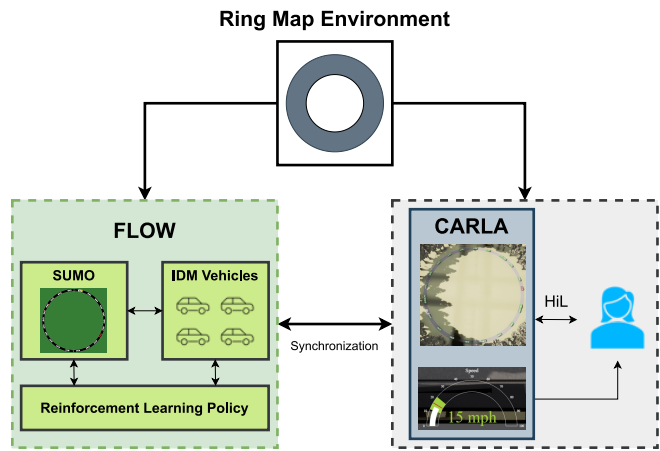

Fig. 1: A block diagram representing the information flow and the different components of our Human-in-the-Loop (HiL) framework.

parameters, they were tested solely in simulation without the involvement of a real human. We would like to extend the simulation framework to include a human-in-the-loop to test the robustness of these policies when combined in a shared control objective.

To this extent, we have developed add-ons to the simulation framework of [8] and propose the use of the CARLA driving simulator to conduct a user study to evaluate piecewise constant policies. In particular, drivers in our shared control experiment are asked to follow the advised action output by the driving policy to reduce overall congestion. In this paper, we present our in-progress work towards this goal and discuss the planned improvements to our experiments.

## II. METHOD

In this section, we describe the different technical components of our shared control framework for co-operative congestion mitigation. Fig. 1 shows an overview of our proposed framework and its different modules.

### A. Piecewise Constant Driving Policies

We now cover a few important features of the congestion mitigation model and policies postulated in [8], but refer the reader to the original paper for more details and proofs. The original setup is based on the Flow framework [4] which is built on top of Simulation of Urban Mobility (SUMO) - a multi-modal, microsimulation package [9]. Flow is a standalone framework for simulating and training RL models for traffic based scenarios.

The simulated environment consists of a single lane circular track with a circumference of 250m with 22 drivers as shown in Fig. 2. We refer to this track as the 'ring' network. This network approximates an infinite highway as there are no incoming or outgoing vehicles. One driver in this scenario is then replaced with an agent that follows the piecewise constant policies generated by the trained RL model as described in [8]. We refer to this agent as the ego vehicle and all other vehicles in the simulation as the non-ego vehicles. Each non-ego vehicle in the network is modeled according to the Intelligent Driver Model (IDM) to simulate human-like behaviour [10]. The simulation of the non-ego vehicles is handled by the SUMO simulator. Fig. 2 shows the ego vehicle in red and the non-ego vehicle in white in the SUMO simulation generated by Flow.

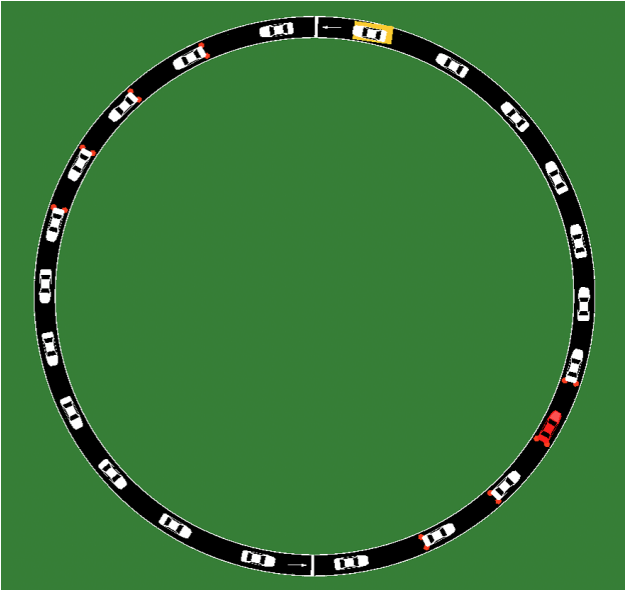

Fig. 2: A birds-eye view of ring network in SUMO.

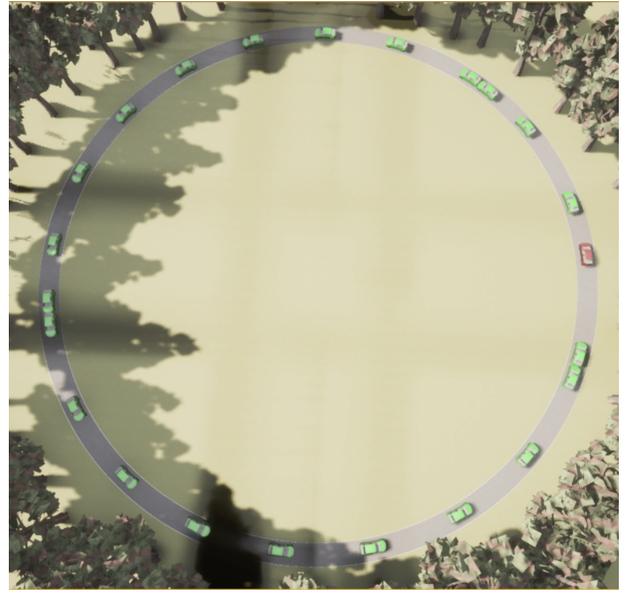

Fig. 3: A birds-eye view of the ring network in the CARLA simulator.

The observation space of the agent is a vector that records its current speed, the speed and the distance to the preceding vehicle, and the circumference of the circular track. With these observations, the policy learns to output the acceleration action that should be held for $\Delta$ timesteps in order to maximize the average speed of all the vehicles on the track and mitigate congestion.

**Acceleration vs. Speed actions:** The authors of [8] chose to output the acceleration action rather than a speed action. This choice was made due to the existence of a trivial speed solution in the simple ring network. However, this is not the case with more complex tracks. This design choice and its limitations are further discussed in Sec. III. For the rest of this paper, we refer to the action of the policy simply as the 'advised action'. Any technical aspect that is discussed below associated with either speed or acceleration would be updated accordingly for uniformity based on the action type.

The RL model is trained with steps that last for 0.1 seconds for a horizon of 8000 steps with 1200 warmup steps. The TRPO algorithm is used to train the model for 500 iterations. A $\Delta$ of 1 timestep was used for training as well as evaluation. The authors make an assumption during training that all advice is followed immediately and exactly. This assumption is relaxed during evaluation by using the IDM model to transition between the current action and the advised action. To go one step further, we propose replacing the IDM model with a human driver. The number of timesteps that the driver takes to perceive and follow the advised action could be recorded as an experimental measurement.

While Flow facilitates the training and evaluation of the RL models, it does not possess an interface for a human driver to test the feasibility of such piecewise constant driving policies. Thus, we propose the use of the CARLA simulator as an interface between the driving policy and a real human.

### B. CARLA Simulation

CARLA is a popular open-source driving simulator developed for autonomous vehicle research [11]. The simulator provides scalability with a server multi-client architecture, a flexible API, and integration for co-simulation with popular traffic simulators such as SUMO.

In order to simulate the ring network, we designed a custom map using RoadRunner [12] and imported the map into CARLA [13]. We used the extensive CARLA python API to build a client program to enable the participants in our study to control the ego vehicle in a world simulating the ring network. We chose to surround the track with an irregular pattern of trees to avoid distracting the users away from the highway ring environment. Fig. 3 shows a birds eye view of the CARLA world with the ego vehicle shown in red and the non-ego vehicles shown in green.

### C. Study Procedure and Setup

In this section, we detail our tentative plan for our proposed user study. We aim to recruit at least 25 participants by advertising the study via email and fliers. Each participant would take part in a 30 minute long session, which is divided into three segments. During the first 7 minutes of the study, the participants are asked to familiarize themselves with the controls and the simulator. Then, they would spend 15 minutes driving in the simulated study environments while following the advice from the driving policy. Finally, the participants would spend 5 minutes answering a questionnaire about their driving experience. The participants would be given a short break between each segment.

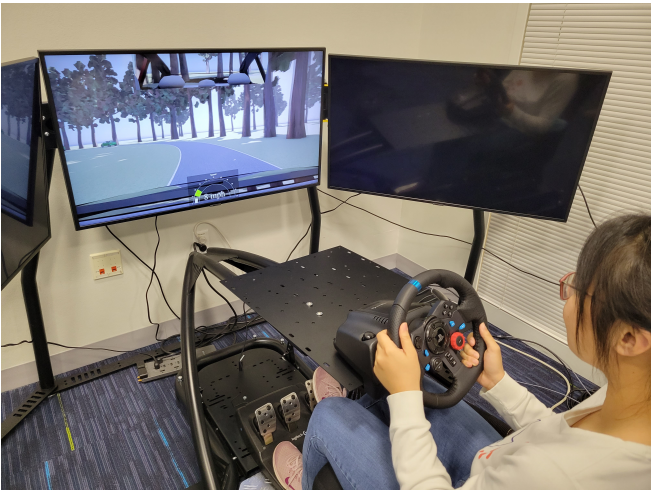

Fig. 4: A participant using the driving simulator to control the ego vehicle

During the first segment, the participants will be given time to familiarize themselves with the controls for the ego vehicle, *i.e.*, the steering wheel, the throttle and brake pedals, and adjust to the user interface. During this initial trial period, the participants will be driving in maps not resembling the testing maps with other non-ego vehicles present.

In the second segment, the participants will be directed to drive in three 5 minute trials on the ring network. They will be asked to follow the action shown on the user interface during each trial. The hyper-parameters, which are discussed later, will be varied across the trials. During each trial, the average speed of all the vehicles will be continually recorded to note the affect on congestion.

Our driving simulator setup is designed to mimic real life driving experiences through the use of the Visaro driving rig and the Logitech G29 Racing Wheel. Fig. 4 depicts a participant driving the ego vehicle using our simulator. To further enhance the simulation experience, a TV setup shows the driver the front view from the vehicle. Fig. 5 shows this front view in more detail. The front view includes an insert of the rear view mirror at the top of the screen to provide a better user experience. The side view mirrors are not included as they would only serve as a source of distraction rather than a utility since the participants are driving on a single lane track.

The speedometer view shown at the bottom of the front view acts as the main user interface between the driving policy and the driver. The speedometer shows the current speed that the user is travelling at visually using a gauge, and also as text for more informed control. The driving policy outputs a singular target value, *i.e.*, the advised action that the ego vehicle should take to mitigate congestion. The actual target value is shown on the speedometer as a red line. In practice, it is hard to maintain one singular action value for a duration of time longer than a couple of seconds due to the nature of the task. Therefore, we define an acceptable range of actions around the target value to show the driver a range of possible actions. This range is indicated in green on the speedometer. Fig. 6 shows examples of the speedometer where the target value is 17 mph with an acceptable range of ±5 mph. Additionally, when the driver is driving within the acceptable range, the text shown at the center of the speedometer changes color to green to provide more feedback to the user. When the user is driving outside the acceptable range, the text is displayed in white. This change can be observed in the two side-by-side images in Fig. 6.

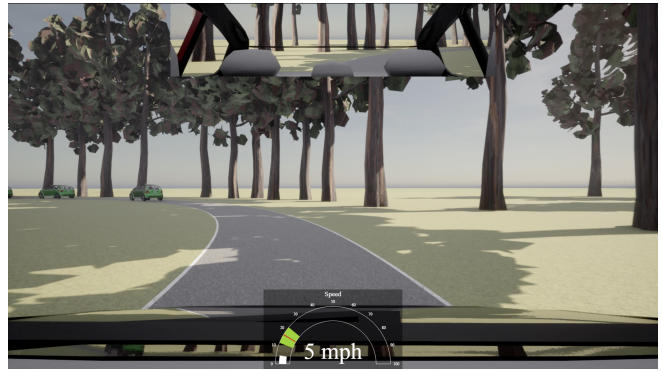

Fig. 5: The front view from the CARLA driver interface

The extent of the range is a hyper-parameter that we would like to vary in our user study. Other hyper-parameters include the density of non-ego vehicles on the track and the IDM parameters for the non-ego vehicles.

**Evaluation:** Currently we propose to evaluate the affect of the human-in-the-loop and the performance of the policies by measuring the average velocity of all the vehicles on the road to measure congestion. A higher average velocity of the vehicles implies that all the vehicles in the network are moving smoothly at a relatively constant speed which indicates that there is less congestion in the network. Likewise, a lower average velocity would indicate that there is likely stop-and-go congestion in the network.

### D. Simulation Synchronization

As discussed in Sec. II-C and highlighted in Fig. 1, the ego vehicle is controlled by the study participant through CARLA, while the non-ego vehicles are controlled by Flow. Hence, the two simulators need to be synchronized so that they share common beliefs about the joint simulation. This synchronization would enable the RL policy model to observe the current state of the ego vehicle and change its advised action. Unfortunately, there is no library to synchronize between Flow and CARLA to provide a seamless user experience. Fortunately, due to the widespread use of the CARLA and SUMO simulators, libraries are available for co-simulation and synchronization between them [14]. Therefore, we were able to develop add-ons to the Flow framework as well as the CARLA-SUMO co-simulation library to achieve our desired synchronization between Flow and CARLA. Thus, any change in either simulation is reflected in the other.

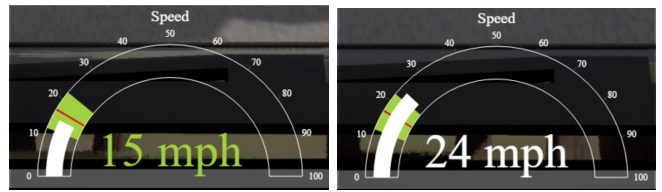

Fig. 6: The visual feedback of when the driver driving within the advised range when compared to when the driver is driving outside the advised range.

In particular, in the CARLA-SUMO co-simulation, we added the ability for CARLA to be able to take in the output of the driving policies and display them on the driver interface. In doing so, we are also able to spawn and mimic the movement of the non-ego vehicles, that are controlled by Flow, in the CARLA environment.

Similarly, for the Flow framework we developed modules to update the state of the ego vehicle in Flow, as seen in CARLA, such that the non-ego vehicles would be reactive to it. As such, we ensure that the non-ego vehicles do not ignore the ego vehicle, as is often the case with co-simulations, thus improving the overall simulation. To this affect, we added functionalities to import custom maps (maps not created in Flow) and allow for a plug-play approach for the environments simulated using Flow.

## III. Limitations and Future Work

In this section, we discuss the current limitations of our work and enumerate areas for improvement.

First, as discussed in Sec. II-A, we would like to attempt to make our system more human-readable by experimenting with both acceleration and speed actions. While a vehicle model can directly utilize acceleration commands to control itself on a road, it may be unintuitive for a user to input throttle commands to follow the given acceleration directions effectively. As such, since all real-world vehicles have a visual speedometer telling the user the current speed of the vehicle, we plan to train our policy to also output speed commands that may be more digestible to human drivers where applicable.

Furthermore, we plan to increase stochasticity in the training process of the policy to make the controller more robust at test time. For example, we can introduce noise to the action input to the vehicle model to copy how humans may not hold a steady throttle command. Another source of randomness can be the simulated roads themselves. Currently, we only train and test our policy on the ring network - a closed, one-lane infinite highway. We hope to apply the policy to a larger set of maps with more complex curves and merging or exit lanes to visualize how robust our model may be on these challenging scenarios. These proposed modifications can decrease the gap when we perform a sim2real transfer to a real-world autonomous vehicle.

Finally, we would like to address how to make our model adaptive to varying human behaviors. Suppose user $A$ drives more conservatively than $B$, or $A$ has a slower reaction time than $B$. A co-operative congestion mitigation policy should be able to tailor directions to each user differently to effectively maximize the reward. In such a task, our policy should, (1) learn the driving style of the human driver, and (2) give directions to maximize reward by outputting actions conditioned on the learnt driving style. Thus, inspired by [15], we plan to learn a latent space of human behaviors at train time, identify a behavior at test time in the latent space, and condition output actions on the identified latent behavior. More specifically, we will learn a latent space of varying reaction times to the action at train time. At test time, we will identify the reaction time of the human in the latent space and generate an action conditioned on the identified behavior. We hope to see that the policy can effectively update its action to minimize congestion from a diverse set of human users.

## IV. Conclusion

In this short paper, we present our in-progress work in developing a framework for studying the utility of piecewise constant driving policies that aide human drivers towards co-operative congestion mitigation in a simulation setting. We hope to conduct our user studies soon and to validate the results obtained in simulation in a human-in-the-loop shared control setting. As discussed in III, our eventual goal is to deploy such a framework in a real world experiment to confirm our findings in the simulated environments.